%% file: main.tex
\pdfoutput=1
\documentclass{article}

\usepackage[final]{venues/colm2026/colm2026_conference}
\usepackage{hyperref}
\usepackage{lineno}
\usepackage{twemojis}     % emoji affiliation markers
\hypersetup{colorlinks=true,citecolor=black,linkcolor=black,urlcolor=black}

\input{manuscript/shared}
\input{manuscript/metadata}

\newcommand{\AffGT}{\twemoji{bee}}          % Georgia Tech
\newcommand{\AffMATS}{\twemoji{fire}}       % MATS Program
\newcommand{\AffGTSafety}{\twemoji{shield}} % Georgia Tech AI Safety Initiative

\title{\PaperTitle}
\author{%
  Mark Riedl$^{\AffGT}$ \enspace \& \enspace Glenn Matlin$^{\AffGT\,\AffGTSafety\,\AffMATS}$ \\[4pt]
  \AffGT\,College of Computing, Georgia Institute of Technology \\
  \AffGTSafety\,Georgia Tech AI Safety Initiative \quad
  \AffMATS\,MATS Program \\[2pt]
  \texttt{riedl@cc.gatech.edu}, \texttt{glenn@gatech.edu}%
}

\begin{document}

\ifcolmsubmission
\linenumbers
\fi

\maketitle
% Camera-ready: workshop header instead of the style's COLM-conference default.
\lhead{Published at the Social Sim'26 Workshop at COLM 2026}
\blfootnote{Equal contribution.}

\input{manuscript/body}

\clearpage
\bibliography{references,manual_sources}
\bibliographystyle{venues/colm2026/colm2026_conference}

\clearpage
\input{manuscript/ethics}

\clearpage
\appendix
\input{manuscript/appendix}

\end{document}

%% file: manuscript/shared.tex
\usepackage[utf8]{inputenc}
\usepackage[T1]{fontenc}
\usepackage{url}
\usepackage{xurl}
\usepackage{booktabs}
\usepackage{tabularx}
\usepackage{longtable}
\usepackage{array}
\usepackage{amsfonts}
\usepackage{amssymb}
\usepackage{microtype}
\usepackage{graphicx}
\graphicspath{{figures/}{./figures/}}
\usepackage{tikz}
\usepackage{enumitem}
\usepackage{xspace}

\usetikzlibrary{arrows.meta,positioning,shapes.multipart,fit}
\setlist[itemize]{leftmargin=1.3em,itemsep=1pt,topsep=2pt}
\setlist[enumerate]{leftmargin=1.5em,itemsep=1pt,topsep=2pt}
\newcolumntype{Y}{>{\raggedright\arraybackslash}X}

\input{macros/main}

\newcommand{\blfootnote}[1]{%
  \begingroup
    \renewcommand{\thefootnote}{}%
    \footnote{#1}%
    \addtocounter{footnote}{-1}%
  \endgroup
}

\ifdefined\IsTwoColumn
  \newenvironment{paperwidetable}[1][t]{\begin{table*}[#1]}{\end{table*}}
\else
  \newenvironment{paperwidetable}[1][t]{\begin{table}[#1]}{\end{table}}
\fi

%% file: macros/main.tex

\input{macros/comments}
\input{macros/references}

%% file: macros/comments.tex
\ifdefined\IsPreprint
  \newcommand{\Mark}[1]{}
  \newcommand{\MarkLeft}[1]{}
  \newcommand{\MarkRight}[1]{}
  \newcommand{\Glenn}[1]{}
  \newcommand{\GlennLeft}[1]{}
  \newcommand{\GlennRight}[1]{}
  \newcommand{\todocite}[1]{}
  \newcommand{\Yixiong}[1]{}
\else
  \newcommand{\Mark}[1]{\textcolor{blue}{[Mark: #1]}}
  \newcommand{\MarkLeft}[1]{\textcolor{blue}{[$\leftarrow$Mark: #1]}}
  \newcommand{\MarkRight}[1]{\textcolor{blue}{[Mark: #1$\rightarrow$]}}
  \newcommand{\Glenn}[1]{\textcolor{orange}{[Glenn: #1]}}
  \newcommand{\GlennLeft}[1]{\textcolor{orange}{[$\leftarrow$Glenn: #1]}}
  \newcommand{\GlennRight}[1]{\textcolor{orange}{[Glenn: #1$\rightarrow$]}}
  \newcommand{\todocite}[1]{\textcolor{red}{[cite #1]}}
  \newcommand{\Yixiong}[1]{\textcolor{brown}{[Yixiong: #1]}}
\fi

%% file: macros/references.tex
\newcommand{\reffig}[1]{Figure~\ref{fig:#1}}
\newcommand{\reftab}[1]{Table~\ref{tab:#1}}
\newcommand{\refsec}[1]{Section~\ref{sec:#1}}

%% file: manuscript/metadata.tex
\newcommand{\PaperTitle}{Position: AI Is Not Ready for Strategic Conflicts}

%% file: manuscript/body.tex
\begin{abstract}
\input{sections/00-abstract}
\end{abstract}

\ifdefined\IsTwoColumn\newpage\fi

\input{sections/01-introduction}

\input{sections/02-decision-influence-not-gameplay}

\input{sections/03-domain-specific-failures}

\input{sections/04-benchmarks-insufficient}

\input{sections/07-implications-for-ml}

%% file: sections/00-abstract.tex
Open-ended strategic wargames are high-stakes LM-based social simulations: they model adversaries, institutions, escalation, plan brittleness, doctrine, and crisis response. Language models (LMs) are attractive because they can play agents, generate scenario branches, adjudicate ambiguous actions, and summarize lessons, but the same affordances make open-ended roles dangerous: model language determines both what an actor attempts and what becomes simulated reality. This position paper argues that no LM-enabled wargame should inform planning, doctrine, policy, or crisis response without an auditable safety case, and that the proper use of open-ended wargames today is to stress-test decision-influencing LM agents. We identify five failure modes: decision laundering, adjudication opacity, role collapse, escalation-through-adjudication, and failure of strategic imagination. Ordinary benchmarks cannot establish safety for these settings. Wargames can expose failures as stress tests; they are not themselves safety cases for consequential use.

%% file: sections/01-introduction.tex
\section{Position and Scope}
\label{sec:intro}

\input{sections/01-introduction/01-pre-table}

%% file: sections/01-introduction/01-pre-table.tex
In 2026, public reporting regarding U.S. defense operations moved language model (LM) decision support from a hypothetical concern to an active policy issue. Reports described Defense Department exploration of Anthropic's Claude model for battlefield simulation alongside sworn declarations detailing xAI's Grok Gov Model integration into Maven Smart Systems for planning, red-teaming, and logistics \citep{stanley_declaration_2026, grenoble_pentagon_2026, fields_pentagon_ai_2026}. While public records leave exact operational roles unverified, these developments signal a rapid institutional trajectory. More broadly, defense disclosures and industry reports describe numerous emerging systems with roles in course-of-action generation, adversary emulation, wargame acceleration, simulation interrogation, or mission-tuned planning support.

Target selection and weapon guidance are not the only high-stakes uses. LMs are increasingly explored for planning, course-of-action selection, and information aggregation---uses ``upstream'' from kinetic or policy effects, shaping choices before leaders decide how to act. Although military wargaming is the primary testbed, these risks apply wherever open-ended simulations inform consequential decisions.

\textbf{Our position is that AI is not ready for strategic conflicts: frontier LMs are entering weapons-adjacent and upstream planning systems without a public, auditable safety case covering role boundaries, uncertainty, adjudication authority, human oversight, and limits on inference; a safety case must be refusal-capable, establishing where systems must not be used, and until such safety cases exist, no LM-enabled wargame or decision-support system should inform planning, doctrine, policy, or crisis response.}\footnote{Studying these systems does not endorse military deployment; we do not condone the use of artificial intelligence in the conduct of war.} The constructive path runs through the wargames themselves: researchers should use open-ended wargames as stress tests that expose how decision-influencing LM agents fail before those failures reach real decisions. When conflict occurs, there will be pressure to use any technology seen as providing a strategic benefit; the time to learn how these systems fail is before that pressure arrives.

The argument proceeds in three steps. \refsec{decision-influence-not-gameplay} shows what separates open-ended wargames from the closed games where LM agents earned their reputation: their product is decision influence, and closed-game performance cannot certify it. \refsec{domain-specific-failures} identifies five failure modes specific to that role: decision laundering, adjudication opacity, role collapse, escalation-through-adjudication, and failure of strategic imagination. \refsec{benchmarks-insufficient} argues that benchmarks and LM-as-judge evaluation cannot establish safety here, and states the norm the field should adopt: no safety case, no high-stakes use.

%% file: sections/02-decision-influence-not-gameplay.tex
\section{Why Open-Ended Wargames Are Different}
\label{sec:decision-influence-not-gameplay}

\input{sections/02-decision-influence-not-gameplay/body}

%% file: sections/02-decision-influence-not-gameplay/body.tex
A serious wargame represents conflict under uncertainty: actors with opposing interests, a synthetic environment, choices, consequences resolved by rules or adjudication, and after-action interpretation intended to inform real-world judgment \citep{perla_what_wargaming_1985, rubel_epistemology_war_2006, us_army_war_college_strategic_wargaming_2015}. Its product is often not a score but a \emph{judgment}: what an adversary might do, which assumptions matter, and which plans are brittle \citep{perla_why_wargaming_2011}. In that sense, a strategic wargame is a social simulation whose output can become part of organizational reasoning.

The safety problem is sharpest when language can shape both sides of the exercise. Fixed games reject invalid moves and resolve consequences through inspectable rules; open-ended language-mediated wargames instead interpret, negotiate, partially accept, or transform novel proposals into scenario state. The danger is practical: the same language interface that makes the exercise flexible can also make unsupported consequences look like simulated facts.

Language models are unusually well-matched to this setting. They can draft plans, maintain role-conditioned dialogue, simulate stakeholders, summarize transcripts, and adjudicate unforeseen actions. Earlier work on tabletop role-playing games, narrative generation, and open-ended text worlds studied similar affordances as precursors to general agent capability \citep{martin_dungeons_dqns_2018, callison-burch_dungeons_dragons_2022, zhu_calypso_llms_2023, cui_mixtureexperts_2024}. The hazard is that an LM hallucination becomes world state, a fluent adjudication becomes evidence, and a polished after-action narrative becomes strategic judgment.

The full pathway is: model-generated action $\rightarrow$ adjudicated consequence $\rightarrow$ accumulated scenario state $\rightarrow$ after-action finding $\rightarrow$ human judgment. A biased adversary model can make a vulnerability appear unimportant; a plausible escalation narrative can make escalation appear inevitable. Similar pathways matter in other LLM-based social simulations, but strategic conflict is the focal use case because wargames sit close to planning, doctrine, crisis reasoning, and escalation.

%% file: sections/03-domain-specific-failures.tex
\section{Failure Pathways}
\label{sec:domain-specific-failures}

\input{sections/03-domain-specific-failures/body}

%% file: sections/03-domain-specific-failures/body.tex
LM safety failures are well known: hallucination, prompt sensitivity, sycophancy, unfaithful reasoning, bias, long-context failures, and brittle out-of-distribution behavior \citep{turpin_language_models_2023, lanham_measuring_faithfulness_2023, liu_lost_middle_2024, sharma_understanding_sycophancy_2024, taubenfeld_systematic_biases_2024, li_llmsasjudges_comprehensive_2024}. While human wargames have long struggled with human bias, opacity, and exercise design constraints \citep{downes-martin_adjudication_diabolus_2013}, LM mediation introduces a distinct structural hazard: *correlated semantic error* across automated roles. In open-ended wargames, generic LM weaknesses combine into domain-specific failure pathways where model outputs dictate simulated reality.

\textbf{Decision laundering.} AI-enabled wargames launder model speculation into strategic judgment. A model-generated branch may begin as a plausible scenario continuation, pass through adjudication, appear in a final summary, and then be treated as a discovered strategic insight. Fluency and narrative confidence are the hazards: a polished after-action narrative makes an outcome feel more causally grounded than the evidence supports, and participants may treat fluent adjudications as neutral exercise products rather than model-generated hypotheses, especially when the narrative matches expectations \citep{ehsan_humancentered_explainable_2020, sharma_why_would_2024}.

\textbf{Adjudication opacity.} The adjudicator is the technical center of the paper. In an open-ended wargame, the adjudicator decides whether a proposed action works, what side effects occur, and what new facts enter the scenario. Human wargaming literature has long treated adjudication as difficult and central \citep{downes-martin_adjudication_diabolus_2013, downes-martin_validity_utility_2017}; LM adjudication makes the difficulty computational and semantic. Players can propose coercive diplomacy, cyber compromise, or alliance manipulation; the system must handle unforeseen actions without over-accepting, refusing too much, or drifting outside the exercise's purpose. The audit objects are assumptions, causal links, alternatives, evidence, uncertainty, and responsibility for major state changes.

\textbf{Role collapse.} Role collapse is the primary delta separating LM-mediated simulations from human wargaming. While human participants bring heterogeneous background knowledge and cognitive diversity, LM-enabled wargames often invite the same model family—or identical base checkpoints—to act as player, adversary, adjudicator, analyst, judge, and summarizer. Role collapse makes a simulation self-confirming: the same latent priors and safety alignment artifacts propose a move, check plausibility, adjudicate consequences, and summarize the lesson. Multi-agent simulations do not solve this when agents share training data, post-training objectives, tools, or prompting conventions; the risk is correlated error across institutional roles. Helpfulness or harmlessness training may distort adversarial roles, softening provocative actions or converging on cooperative behavior when the exercise needs a capable opponent \citep{askell_general_language_2021, sharma_understanding_sycophancy_2024, yi_too_good_bad_2025}. Omitted context becomes a hidden prior: the model fills evidentiary gaps with pattern completion while presenting the result as strategic reasoning.

\textbf{Escalation-through-adjudication.} Escalation stems from player choices and adjudicator interpretations. If a model repeatedly narrates adversaries as interpreting ambiguity in the most hostile way, the exercise manufactures escalation pressure; prior work shows LMs display escalatory tendencies in military and diplomatic decision-making \citep{rivera_escalation_risks_2024}. The mechanism is path-dependent: an early mistaken assumption about intent, logistics, or feasibility can persist as scenario state and influence later turns. The true failure is lost strategic continuity: an adjudicator's early framing makes escalation appear likely, rational, or unavoidable even when no player chose it.

\textbf{Failure of strategic imagination.} Wargames at their best surface branches planners had not anticipated, and a capable adversary explores options a defender did not imagine. LM-based agents interpolate within recorded strategies rather than extrapolating beyond them. On the adversary side, the result is an opponent that argues plausibly within the player's framing but does not stress-test a plan from novel angles. On the planner side, the same interpolation suppresses unexpected options. The aspirational use of an AI participant is to be at least as creative as human participants; the realized capability is often closer to a fluent restatement of strategy literature. That gap is a safety property that should be measured rather than assumed.

These failures compound. Adjudication opacity makes decision laundering more likely because fluent adjudications carry unsupported assumptions into final summaries. Role collapse amplifies that hazard: the same latent biases can produce both the move and the adjudication that justifies it. Escalation-through-adjudication compounds as early framing becomes durable simulated reality. Failure of strategic imagination produces an exercise that is internally fluent but strategically narrow. The safety question is therefore not, ``Did the agent win?'' It is, ``Which generated claims entered the simulated world, why were they accepted, and what real decisions might they influence?''

%% file: sections/04-benchmarks-insufficient.tex
\section{Benchmarks Are Not Safety Cases}
\label{sec:benchmarks-insufficient}

\input{sections/04-benchmarks-insufficient/01-body}

\input{tables/tab-eval-vs-safetycase}

\input{tables/tab-safetycase-compact}

\input{sections/04-benchmarks-insufficient/02-after-table}

%% file: sections/04-benchmarks-insufficient/01-body.tex
Benchmarks remain useful intermediate steps, but they are not safety cases. Three limits matter for open-ended wargames \citep{kapoor2024agents}: benchmark over-optimization, distributional mismatch between fixed tasks and adaptive strategic interaction, and speculative futures with no ground-truth label. Quality therefore requires expert review of exercise traces.

Common substitutes have the same limit. LLM-as-judge evaluation scales review but introduces prompt sensitivity and correlated errors \citep{li_llmsasjudges_comprehensive_2024}. As operationalized in \reftab{eval-vs-safetycase}, standard benchmarks and ad-hoc red-teaming identify specific failure modes but fall short of establishing auditable operational boundaries. Human-likeness is also insufficient: a safe adversary model must be calibrated, role-faithful, non-sycophantic, and clear about uncertainty. The relevant audit object is often the adjudication or summary trace: why a contested action was accepted, what assumptions linked action to consequence, what alternatives were considered, and how uncertainty was represented. Model-level interpretability may help inspect role leakage, but cannot replace exercise-level artifact capture.

A \emph{safety case} is a structured argument, supported by retained evidence, that a system is acceptably safe for a specified use \citep{kelly1998arguing}, recently adapted to advanced AI systems by \citet{clymer2024safetycases}. It states the use, hazards, controls, supporting evidence, counter-evidence, and limits on inference. Because language spaces lack numeric failure probabilities, safety cases must specify refusal bounds where deployment is prohibited. For open-ended wargames, the central norm is simple: \textbf{No safety case, no high-stakes use.}

%% file: tables/tab-eval-vs-safetycase.tex
\begin{table}[t]
\caption{Operational comparison of evaluation paradigms for high-stakes LM decision support.}
\label{tab:eval-vs-safetycase}
\centering
\small
\setlength{\tabcolsep}{4pt}
\renewcommand{\arraystretch}{1.1}
\begin{tabularx}{\linewidth}{>{\raggedright\arraybackslash}p{2.2cm}>{\raggedright\arraybackslash}X>{\raggedright\arraybackslash}X}
\toprule
\textbf{Paradigm} & \textbf{Operational Scope} & \textbf{Strategic Limitation} \\
\midrule
\textbf{Benchmark Testing} & Static task accuracy, multi-choice QA, or closed-game win rates. & Measures raw capability; cannot evaluate open-ended scenario drift or decision influence. \\
\textbf{Red-Teaming / Human Review} & Point-in-time vulnerability discovery and expert inspection. & Identifies specific failure cases but lacks structured claim-evidence mapping or operational bounds. \\
\textbf{Auditable Safety Case} & Structured safety claims, auditable adjudication traces, refusal bounds. & Requires ongoing maintenance; explicitly defines where systems must not be used. \\
\bottomrule
\end{tabularx}
\end{table}

%% file: tables/tab-safetycase-compact.tex
\begin{table}[t]
\caption{Minimum retained evidence for high-stakes LM-enabled open-ended wargame safety cases.}
\label{tab:safetycase-compact}
\centering
\small
\setlength{\tabcolsep}{4pt}
\renewcommand{\arraystretch}{1.05}
\begin{tabularx}{\linewidth}{>{\raggedright\arraybackslash}p{2.9cm}Y}
\toprule
\textbf{Component} & \textbf{Minimum retained evidence} \\
\midrule
Scope and use & Named decision context, intended users, forbidden uses, and downstream distribution path \\
Role separation & Role map, model and prompt assignments, and independent review of high-impact claims \\
Adjudication trace & Assumptions, sources, alternatives, uncertainty, and responsibility for major state changes \\
Robustness & Paraphrase, prompt-sensitivity, cross-model, and adversarial tests \\
Human contestability & Named intervention and override points, expert review, and preserved disagreements \\
Summary discipline & Trace support for consequential after-action claims and calibrated confidence language \\
\bottomrule
\end{tabularx}
\end{table}

%% file: sections/04-benchmarks-insufficient/02-after-table.tex
A safety case does not certify that a wargame is ``true.'' It documents whether a particular LM-enabled exercise is fit for a particular decision-support role, and it should make non-use available. Evidentiary requirements scale with consequence; a safety case is a precondition, not a deployment license.

%% file: sections/07-implications-for-ml.tex
\section{Implications and Conclusion}
\label{sec:implications-for-ml}

\input{sections/07-implications-for-ml/body}

%% file: sections/07-implications-for-ml/body.tex
Three research directions follow: \textbf{adjudication interpretability} should examine which causal assumptions, sources, and role commitments support major adjudications; \textbf{open-action robustness} should test semantically novel moves, paraphrases, hybrid-domain actions, and attempts to exploit ambiguity \citep{peng_detecting_adapting_2021}; and \textbf{summary-effect evaluation} should measure how summaries, confidence language, and narrative framing affect human interpretation.

Negative findings help define where a system should not be trusted: role drift, unsupported consequences, collapsed dissent, escalation bias, and brittle behavior.

The norm has addressable audiences: procurement offices can require a safety case before an LM-enabled exercise informs doctrine or planning; reviewers can require exercise-trace artifacts and disclosed role boundaries from wargaming papers; labs shipping government-facing models can publish role-boundary and escalation evaluations. While motivated by strategic conflict, these evaluation norms apply across high-stakes social simulations broadly whenever open-ended LMs generate scenarios, adjudicate outcomes, or synthesize policy summaries.

In open-ended wargames, the hardest part is not winning the game; it is knowing when the game should not be trusted.

%% file: manuscript/ethics.tex
\section*{Ethics Statement}

This position paper addresses a high-risk dual-use setting: language-model support for strategic wargaming, planning, doctrine, policy, and crisis response. The paper does not introduce a deployable system, release operational scenarios, provide instructions for military use, or report experiments with human subjects. Its purpose is risk assessment: to argue that open-ended wargames can stress-test decision-influencing language-model agents, and that high-stakes use requires auditable safety cases. We do not endorse military deployment of language models; we argue that, absent retained evidence and contestable review, non-use may be the appropriate conclusion.

\section*{Use of Generative AI Tools}

The authors used language-model assistance during manuscript preparation to compress existing material, identify consistency issues, and prepare submission-support materials. The authors reviewed and are responsible for all claims, citations, and final wording. No language model was used to generate empirical data, figures, evaluation results, or new bibliography entries for this submission.

%% file: manuscript/appendix.tex
\input{sections/13-app-public-evidence}

\input{sections/10-app-failure-pathways}

\input{sections/11-app-evaluation-safetycase}

\input{sections/12-app-objections}

%% file: sections/13-app-public-evidence.tex
\section{Public Evidence: Systems Entering Planning and Wargaming}
\label{sec:app-public-evidence}

Public evidence supports a narrow claim: AI and LM systems are being researched, procured, tested, or advertised for roles in planning, course-of-action generation, simulation, wargaming, training, and after-action workflows. A public source is not treated as evidence of operational reliance unless it says so directly.

\subsection{2026 source trail}

The cited 2026 record has three parts. First, several news outlets reported that the US Department of Defense used a version of Anthropic's Claude model in connection with target selection and battlefield simulations.\footnote{Independently reported by \href{https://www.theguardian.com/technology/2026/mar/01/claude-anthropic-iran-strikes-us-military}{The Guardian}, \href{https://www.washingtonpost.com/technology/2026/03/04/anthropic-ai-iran-campaign/}{The Washington Post}, \href{https://www.cbsnews.com/news/anthropic-claude-ai-iran-war-u-s/}{CBS News}, \href{https://www.nbcnews.com/tech/tech-news/us-military-using-ai-help-plan-iran-air-attacks-sources-say-lawmakers-rcna262150}{NBC News}, and \href{https://thehill.com/policy/defense/5799136-claude-pentagon-iran-war/}{The Hill} (Feb.--Mar.\ 2026; links embedded).} Second, a June sworn declaration described xAI's Grok Gov Model in Maven Smart Systems for targeting, intelligence, planning, red-teaming, logistics, and sustainment, and stated that MSS workflows enabled U.S. forces to deploy over 2,000 munitions to 2,000 targets in 96 hours during Operation Epic Fury \citep{stanley_declaration_2026}. Third, Yahoo News and The Hill reported the filing; Yahoo noted that the public record did not establish Grok's exact role in launches or target selection \citep{grenoble_pentagon_2026,fields_pentagon_ai_2026}.

The inventory below records named systems, source-supported role descriptions, sources, and claim boundaries. It separates research prototypes, procurement or vendor claims, training and exercise use, and operational reliance when the public record supports those distinctions.

\subsection{Country/entity summary}
\label{sec:app-systems-summary}

\reffig{systems-coverage} maps the 26 publicly documented examples by country/entity group and primary pipeline role; \reftab{systems} summarizes the same inventory by country or entity group. The full row-by-row inventory is in \reftab{systems-full}.

\input{figures/fig-systems-coverage}

\input{tables/tab-systems}

\subsection{Full public-system inventory}
\label{sec:app-full-system-inventory}

\input{tables/tab-systems-full}

\subsection{Public-system search protocol}
\label{sec:app-system-search}

The inventory in \reftab{systems-full} includes only named systems, projects, exercises, or programs for which a public source supports an AI, LM, agent, RAG, generative-AI, or reinforcement-learning role in wargaming, planning, course-of-action analysis, adjudication support, training, or an after-action pipeline. Known examples such as JHU/APL GenWar Sim, DIU Thunderforge, Scale AI Donovan in the USARNORTH wargame, Hadean populAI/dominAI, Dstl/Frazer-Nash analysis of \emph{Command: Modern Operations} outputs, and the U.S. Army CGSC AI-enabled PME wargame were used to set this boundary before adding international examples.

Sources were prioritized in this order: official government, military, defense-ministry, service, laboratory, academy, defense-innovation, procurement, patent, and conference materials; military-academic journals and government-funded research; then vendor pages when they named the system, customer, exercise, program, or contract. News sources were used mainly to identify names and search terms. For each included row, the review recorded the system name, institution, source type, source-supported AI role, status of use, claim boundary, and safety relevance.

The claim-boundary rule was conservative: source evidence for ``AI-assisted planning'' was not treated as evidence of operational reliance; simulator- or human-adjudicated outcomes were not described as standalone AI adjudication; educational examples were classified as PME/training exposure; procurement, SBIR, patent, and vendor pages were classified as proposed, prototype, or vendor capability unless public evidence showed exercise use or deployment; and platform autonomy, ISR, targeting, or kill-chain automation was excluded unless tied to planning, simulation, wargaming, or COA analysis.

%% file: figures/fig-systems-coverage.tex
\begin{figure}[!ht]
\centering
\includegraphics[width=\linewidth]{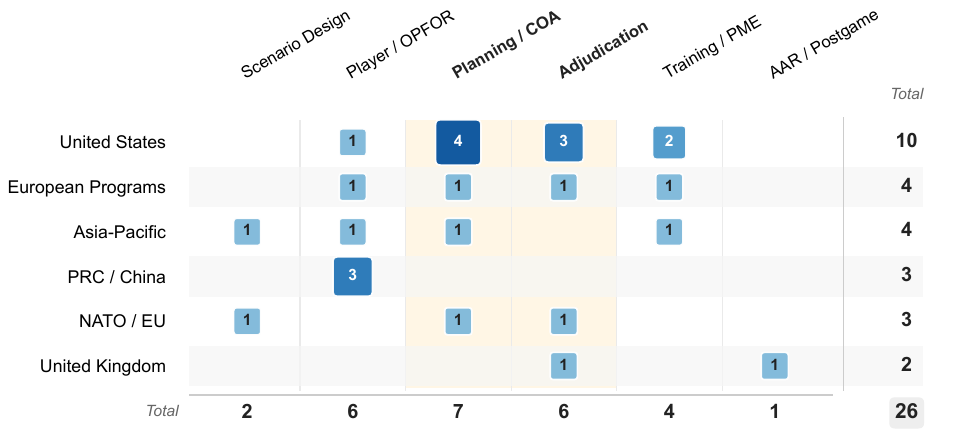}
\caption{26 publicly documented AI/LM systems for open-ended wargaming, by country/entity group (rows) and primary pipeline role (columns). Marginal totals are shown on the right and bottom. Full names, sources, and claim boundaries are in \reftab{systems-full}.}
\label{fig:systems-coverage}
\end{figure}

%% file: tables/tab-systems.tex
\begin{table}[t]
\caption{Country/entity-group summary of the 26 publicly documented systems in the inventory. Compound rows count once; full system names, source-supported roles, sources, and claim boundaries are in \reftab{systems-full}.}
\label{tab:systems}
\centering
\scriptsize
\setlength{\tabcolsep}{3pt}
\renewcommand{\arraystretch}{1.08}
\begin{tabularx}{\linewidth}{>{\raggedright\arraybackslash}p{2.45cm}>{\centering\arraybackslash}p{0.62cm}YY}
\toprule
\textbf{Country/entity group} & \textbf{Rows} & \textbf{Representative systems} & \textbf{Roles represented} \\
\midrule
United States & 10 & GenWar, WarMatrix, Thunderforge, SCEPTER, Donovan, USAWC/CGSC & Planning/COA advice, adjudication/simulation coupling, PME/training, AAR, agent research \\
United Kingdom & 2 & Hadean populAI/dominAI, Dstl CMO pipeline & Simulation coupling, COA generation, AAR/postgame analysis \\
PRC / China & 3 & MiaoSuan, MaCA/Zhanlu/Mozi, Self-Generated Wargame AI & Player/OPFOR agents, agent competitions, COA and exercise evaluation \\
NATO / EU institutions & 3 & NATO TIDE WarSim, HEDI, BortChat & Prototype wargaming, scenario analytics, COA advice, decision articulation \\
European national or multinational programs & 4 & SWORD/SOULT, GhostPlay, Baltic Shadow, FOI COA generation & Training, agent simulation, scenario control, COA search \\
Asia-Pacific national programs & 4 & ChangJo21/NORAM, ATLA, WARDEC, NWS-980 & OPFOR automation, scenario design, planning support, training \\
\bottomrule
\end{tabularx}
\end{table}

%% file: tables/tab-systems-full.tex
\begingroup
\scriptsize
\setlength{\tabcolsep}{3pt}
\renewcommand{\arraystretch}{1.03}
\begin{longtable}{>{\raggedright\arraybackslash}p{2.75cm}>{\raggedright\arraybackslash}p{5.15cm}>{\raggedright\arraybackslash}p{5.25cm}}
\caption{Publicly documented AI/LM-enabled systems, programs, and research prototypes relevant to open-ended military wargaming and planning. The inventory records safety relevance and public-evidence boundaries rather than treating every row as validated deployment.}
\label{tab:systems-full}\\
\toprule
\textbf{System (institution)} & \textbf{What it does} & \textbf{Safety relevance and claim boundary} \\
\midrule
\endfirsthead
\toprule
\textbf{System (institution)} & \textbf{What it does} & \textbf{Safety relevance and claim boundary} \\
\midrule
\endhead

\textbf{GenWar TTX / GenWar Sim} \emph{(JHU/APL)}
& Integrates LLMs with AFSIM for natural-language interaction, COA exploration, AI-informed dialogue, and physics-based adjudication.
& Safety relevance is the pipeline: free-text moves can pass through model-mediated exploration, physics-based adjudication, and exercise evidence. The row does not claim standalone LM adjudication. \\

\textbf{WarMatrix} \emph{(Department of the Air Force)}
& AI-enabled wargaming environment used in the GE 26 Benchmark Wargame; integrates models, data, workflows, scenario development, and simulation-informed adjudication.
& Places AI inside a live/semi-live wargame workflow; traceable outputs and postgame datasets can shape force-design and planning judgments. Public evidence supports exercise use, not the exact authority of model outputs. \\

\textbf{Thunderforge} \emph{(DoD / Defense Innovation Unit)}
& Prototype AI planning ecosystem intended to combine LLMs, AI-driven simulations, and interactive agent-based wargaming for operational/theater planning and COA refinement.
& Relevant because it targets planning infrastructure that combines LM planning, simulation, and agentic wargaming. Public sources support prototype-contract scope and stated deployment intent, not independent evidence of operational reliance. \\

\textbf{SCEPTER / STRATEGIC Engine} \emph{(DARPA / Code Metal)}
& Machine-speed COA generation and simulator validation under SCEPTER; SBIR platform scope adds LLM/RL-enabled distributed wargaming, planning, adjudication, BDA, and automated COA comparison.
& High-speed option generation and simulator validation can narrow what human planners review. Public STRATEGIC Engine evidence is an SBIR scope, not validated fielded performance. \\

\textbf{COA-GPT} \emph{(Goecks \& Waytowich)}
& LLM algorithm for doctrinal COA generation and refinement from text/image mission inputs; evaluated in a militarized StarCraft II scenario.
& Relevant as doctrinal LM planning assistance and rapid COA generation in a militarized testbed. It is a research prototype, not evidence of operational planning infrastructure. \\

\textbf{Donovan in USARNORTH wargame} \emph{(US Army / Scale AI)}
& Scale AI GenAI system tested in a classified USARNORTH theater-Army wargame; the public account centers on an adviser role while discussing scenario developer, adjudicator, and designer as possible broader roles.
& Concrete military exercise use and a plausible role-expansion path; public evidence centers on an adviser role and should not be read as proof that one LM performed all wargame roles. \\

\textbf{Hadean populAI / dominAI} \emph{(Hadean)}
& Vendor stack for wargaming, synthetic environments, BLUFOR/OPFOR COA generation, parallel simulation, speedups, and AAR/PXR support.
& Shows vendor integration of COA generation, simulation, and AAR/PXR support. Evidence is vendor capability language, not independent validation of every claimed function. \\

\textbf{Dstl CMO pipeline} \emph{(UK MoD / Frazer-Nash)}
& Local LLM/RAG pipeline over \emph{Command: Modern Operations} scenario outputs for postgame interrogation, summarization, and dissemination.
& Shows laundering risk through postgame interrogation and summaries even without live LM gameplay or adjudication. Public evidence supports a local analysis pipeline over CMO outputs. \\

\textbf{USAWC \emph{Pacific Strategy}} \emph{(US Army War College)}
& PME Free-Kriegsspiel-style exercise in which LLMs supported intelligence summaries, student plan drafting, faculty adjudication support, timelines, O\&I reports, and BDA generation under human review.
& Direct example of LM assistance around open-ended adjudication and after-action products. The boundary is PME/training under human review, not operational decision support. \\

\textbf{CGSC AI-enabled wargame} \emph{(US Army CGSC)}
& PME classroom wargame in which students used AI-enabled tools to explore, review, and refine COAs over more turns after AI-use instruction and human-override guidance.
& Shows AI-shaped COA exploration inside staff education and possible training-induced trust in polished outputs. Boundary is classroom use with instruction and human override, not operational use. \\

\textbf{Snow Globe} \emph{(IQT Labs)}
& Open-source LLM multi-agent architecture for qualitative open-ended wargames; scenario preparation, play, adjudication/analysis, and postgame products can be AI, human, or hybrid.
& Directly instantiates the paper's hazard class: language-generated moves, role play, adjudication/analysis, and summaries in one open-ended loop. Boundary is open-source research system. \\

\textbf{ERDC AI-enabled wargaming agent training} \emph{(U.S. Army ERDC)}
& Trains AI agents for wargaming using hierarchical reinforcement learning and abstraction methods for combat modeling and simulation.
& Relevant to the closed-modeling side: agent abstractions and reward structures can shape what behavior appears plausible. Evidence is agent-training research, not LM-mediated operational use. \\

\textbf{MiaoSuan Wargame} \emph{(Chinese academic researchers)}
& Multi-mode imperfect-information wargaming platform for AI-agent experimentation.
& Provides a public research substrate for training and comparing wargame agents. Evidence supports an academic platform, not public proof of operational PLA use. \\

\textbf{NATO TIDE WarSim / Wargaming LLM prototype} \emph{(NATO TIDE Hackathon)}
& Hackathon prototype using an LLM-driven wargaming simulator and related NATO TIDE work on LLM support for wargaming.
& Shows rapid prototyping of LM-mediated wargame workflows for mission users. Hackathon evidence supports prototype capability, not validation. \\

\textbf{SWORD / SOULT} \emph{(MASA / French Army)}
& Constructive simulation used for staff training, planning support, operational research, C2 stimulation, and AAR; MASA describes behavioral decision AI and R\&D for automated COA evaluation and causal AAR graphs.
& Fielded training/planning simulation is relevant because automation can enter COA evaluation and AAR products. Public COA/AAR automation claims remain R\&D boundaries, not proof of operational reliance on AI-selected plans. \\

\textbf{ChangJo21 / NORAM AI-CGF} \emph{(ROK Army / ROK Navy researchers)}
& Uses named ROK wargame models for deep-learning OPFOR automation and evolutionary scenario generation for naval computer-generated forces.
& Connects AI to training and plan-analysis substrates through existing wargame models. Evidence supports research prototypes, not fielded autonomous staff work. \\

\textbf{MaCA, Zhanlu, and Mozi} \emph{(PRC public wargaming ecosystem)}
& Public or semi-public PRC sources describe multi-agent combat arenas, AI wargame competitions, human-machine confrontation systems, and Mozi's use for operational-concept exploration and exercise/COA evaluation.
& Important international evidence of AI wargaming activity. Much of the record is competition, secondary analysis, or closed-version description, not independently verified operational use. \\

\textbf{Self-Generated Wargame AI / battle-plan-driven agents} \emph{(PRC academic researchers)}
& LLM, imitation-learning, and reinforcement-learning papers test wargame agents that plan tasks, issue natural-language commands, or learn actions from battle plans.
& Relevant to LMs and learned agents as wargame players and planners. The safety claim is about prototype decision behavior, not deployment. \\

\textbf{ATLA scenario support / AI staff} \emph{(Japan MoD / Fujitsu)}
& Official Japanese projects cover AI-assisted exercise-scenario creation, semi-automated staff operational analysis, and multi-AI-agent staff support for decision acceleration.
& Moves AI into exercise design and staff-work pipelines. Public evidence establishes project and research scope, not operational dependence. \\

\textbf{GhostPlay / GhostPlay@SEA} \emph{(Bundeswehr-funded consortium)}
& Synthetic battlefield and maritime extension for AI-controlled red/blue avatars, tactical AI, red-teaming, and testing AI-supported military systems.
& Relevant for simulation-coupled tactical AI, red-team realism, and testing AI-supported military systems. Boundary is R\&D/evaluation, not live command authority. \\

\textbf{WARDEC} \emph{(Indian Army)}
& Wargaming Development Centre ecosystem for operational planning, training, doctrinal innovation, AI/ML/VR/AR integration, and decision-support tools such as automated IPB and combat decision resolution.
& Formalizes AI-adjacent wargaming and decision-support infrastructure. Official evidence supports simulation-enabled decision-support tools; richer AI-design claims remain secondary. \\

\textbf{Baltic Shadow} \emph{(Netherlands Defence Academy)}
& PME wargame in which instructors and experts use generative-AI input to keep a cyber and information-operations scenario dynamic and challenging.
& Educational-use case in which generative-AI input shapes scenario control and feedback. Humans remain the instructors and adjudicators. \\

\textbf{Autonomous COA generation} \emph{(FOI / Swedish researchers)}
& FOI work generates and evaluates thousands of mechanized-battalion COA alternatives for a decision-maker within a sequential decision-making framework.
& Shows AI advice and COA search entering exploratory wargaming. Boundary is controlled experiment/decision-support research, with risk of option steering and apparent tactical authority. \\

\textbf{SAS-172 Chatbot / BortChat} \emph{(NATO MW COE)}
& Generative-AI/LLM chatbot support in a 2025 mountain-warfare MDO wargame for situational analysis, commander's intent, alternative COAs, and decision articulation.
& Experimental wargame evidence that LMs can alter tempo and framing of tactical choices. Not evidence of operational deployment. \\

\textbf{NWS-980 AI integration} \emph{(Royal Thai Navy)}
& Planned incorporation of LLMs and agentic AI into the Naval Warfare Simulator to train students and help staff respond to student actions.
& Shows diffusion of LLM/agentic AI into naval training plans beyond major AI-defense powers. Boundary is explicitly early-stage training integration. \\

\textbf{HEDI AI in M\&S-enabled wargaming} \emph{(European Defence Agency)}
& Proof-of-concept procurement for integrating AI into modeling-and-simulation-enabled wargaming, including adaptive threat scenarios, real-time analysis, feedback, and performance metrics.
& Procurement scope shows institutional demand for AI-mediated wargaming analytics. Treat as planned proof of concept, not validated deployment. \\

\bottomrule
\end{longtable}
\vspace{2pt}
\noindent\begin{minipage}{0.98\linewidth}
\emph{Abbreviations:} COA = course of action; PME = professional military education; AAR/PXR = after-action/post-exercise report; BDA = battle-damage assessment; O\&I = operations and intelligence.

\emph{Sources:} GenWar \citep{jhuaplGenWarSim2025,jhuaplGenerativeWargaming2025}; WarMatrix \citep{usaf2026warmatrix}; Thunderforge \citep{diuThunderforge2025}; SCEPTER/STRATEGIC Engine \citep{darpa_scepter,sbir2025strategicengine}; COA-GPT \citep{goecks_generative_2024}; Donovan \citep{barryWilcox2025centaur}; Hadean \citep{hadeanWargaming2026}; Dstl/Frazer-Nash \citep{dstlFrazerNash2025}; USAWC \emph{Pacific Strategy} \citep{spahr2025usawcai}; CGSC \citep{armyCGSC2026aiwargame}; Snow Globe \citep{hogan_openended_wargames_2024}; ERDC \citep{rinaudo_artificial_intelligence_2024}; MiaoSuan \citep{xu_miaosuan_wargame_2022}; NATO TIDE \citep{softserve2024tidewargamingllm,warsim2024tide}; SWORD/SOULT \citep{masaSwordSoult2026,masaAiCommand2023}; ROK ChangJo21/NORAM \citep{lee2021changjo21Opfor,kim2022noramAiCgf}; PRC MaCA/Zhanlu/Mozi \citep{rocAirForce2024AiWargame,c2Society2020Maca}; PRC academic wargame agents \citep{sun_self_generated_2023,sun2024battlePlanWargameAgent}; Japan ATLA/Fujitsu \citep{atlaRapidScenarioAi2023,atlaDecisionAcceleration2025,fujitsuDefenseMultiAiAgent2025}; GhostPlay \citep{hensoldtGhostPlay2021,ghostplayProject2026,twentyoneStrategiesGhostPlaySea2025}; WARDEC \citep{pibWardec2026}; Baltic Shadow \citep{defensieBalticShadow2025}; autonomous COA generation \citep{schubert2025autonomousCoa}; NATO MW COE \citep{mwcoeBortChat2025}; NWS-980 \citep{usniRtnNws9802026}; EDA HEDI \citep{edaHediAiWargaming2024}.
\end{minipage}
\endgroup

%% file: sections/10-app-failure-pathways.tex
\section{Expanded Failure Pathways}
\label{sec:app-failure-pathways}

\input{sections/10-app-failure-pathways/original-failure-pathways}

%% file: sections/10-app-failure-pathways/original-failure-pathways.tex
LM safety failures are well known: hallucination, prompt sensitivity, sycophancy, unfaithful reasoning, bias, long-context failures, and brittle out-of-distribution behavior \citep{turpin_language_models_2023, lanham_measuring_faithfulness_2023, liu_lost_middle_2024, sharma_understanding_sycophancy_2024, taubenfeld_systematic_biases_2024, li_llmsasjudges_comprehensive_2024}. In open-ended wargames, these generic failures combine into domain-specific pathways; \citet{hogan_openended_wargames_2024} have begun to document these pathways empirically in open-ended LLM wargames. The problem is not only that an LM might be wrong. The problem is that an LM might be wrong while playing a role that determines what the exercise says about the world and what humans later believe about it.

\textbf{Decision laundering.} AI-enabled wargames launder model speculation into strategic judgment. A model-generated branch may begin as a plausible scenario continuation, pass through an adjudication step, appear in a final summary, and then be treated by decision-makers as a discovered insight. Fluency is the hazard: LMs are trained to produce coherent language, and recent work suggests increasing capability in persuasion and forecasting-like judgment \citep{karger_forecastbench_dynamic_2025, schoenegger_large_language_2025}. A polished after-action narrative makes an outcome feel more causally grounded than the evidence supports, and participants treat fluent adjudications as neutral exercise products rather than model-generated hypotheses---especially when the narrative matches their expectations \citep{ehsan_humancentered_explainable_2020, sharma_why_would_2024}.

Serious wargames are valuable because they reveal assumptions and elicit expert judgment \citep{perla_why_wargaming_2011, reddie_nextgeneration_wargames_2018}; if the model supplies undisclosed assumptions, the exercise converts ungrounded generation into institutional belief, compounding judgment errors over long horizons.

\textbf{Adjudication opacity.} We argue that the adjudicator is a high-leverage component in many open-ended wargames: it decides whether a proposed action works, what side effects occur, and what new facts enter the scenario. Human wargaming literature has long treated adjudication as difficult and central \citep{downes-martin_adjudication_diabolus_2013, downes-martin_validity_utility_2017}; LM adjudication makes the difficulty computational and the action space semantic. Players can propose coercive diplomacy, cyber compromise, legal maneuver, financial pressure, deception, alliance manipulation, humanitarian relief, or hybrid moves that cross domains; the system must handle unforeseen actions without over-accepting, refusing too much, or drifting outside the exercise's purpose. A fluent adjudication can hide unsupported causal assumptions, biased analogies, or inconsistent standards, and explanations do not solve the problem if they are themselves unfaithful or post hoc \citep{turpin_language_models_2023, lanham_measuring_faithfulness_2023}.

\textbf{Role collapse.} LM-enabled wargames often invite the same model family to act as player, adversary, adjudicator, analyst, and summarizer. Role collapse makes a simulation self-confirming: the same latent biases could propose a move, check its plausibility, and summarize its lesson. Multi-agent simulations do not solve this when agents share training data, post-training objectives, tools, or prompting conventions; the risk is correlated mistakes across institutional roles, not just collusion. Post-training for helpfulness and harmlessness compounds the problem when the role expected by the exercise is adversarial: models tuned to suppress harmful or non-normative outputs and prefer cooperative framings can make poor antagonists, soften provocative actions, and converge on helpful behavior even when explicitly cast as villains \citep{peng_reducing_nonnormative_2020, askell_general_language_2021, sharma_understanding_sycophancy_2024, yi_too_good_bad_2025}. The data side compounds the problem in a different direction. Strategic conflict involves doctrine, logistics, domestic politics, law, organizational culture, classified intelligence, and tacit subject-matter-expert (SME) judgment---much of it local, current, private, or contested. Omitted context becomes a hidden prior: the model fills evidentiary gaps with pattern completion while presenting the result as strategic reasoning.

\textbf{Escalation-through-adjudication.} Escalation is not only a player choice---it can be introduced by the adjudicator's interpretation of consequences. If a model repeatedly narrates adversaries as interpreting ambiguity in the most hostile way, the exercise manufactures escalation pressure; prior work shows LMs display escalatory tendencies in military and diplomatic decision-making \citep{rivera_escalation_risks_2024}. The mechanism is path-dependent. Wargames accumulate state, and an early error---a mistaken assumption about actor intent, logistics, or technical feasibility---propagates across many turns and later appears as context. Long-context models still exhibit failures in using information across long contexts \citep{liu_lost_middle_2024, modarressi_nolima_longcontext_2025}; in open-ended wargames, the failure is not forgetting a fact but losing strategic continuity, so an adjudicator's early framing makes escalation appear likely, rational, or unavoidable even when no player chose an escalatory action.

\textbf{Failure of strategic imagination.} Wargames at their best surface options that planners had not considered, and a capable adversary is dangerous because it explores branches a defender did not anticipate. LM-based agents are calibrated to interpolate within the strategies humans have already written down, not to extrapolate beyond them; the same training-distribution gap manifests on both sides of the table. On the adversary side, it appears as \emph{counterfactual flatness}: closed-game agents like AlphaZero rely on tree search and self-play to surface non-obvious continuations \citep{silver_mastering_game_2017}, while LM-based adversaries typically produce greedy single-shot completions, with no equivalent of large-scale Monte Carlo Tree Search and only limited self-play; inference-time search and RL-tuned reasoning models partially close this gap in agent settings, but deployed open-ended wargame stacks rarely use either. The result is an adversary that argues plausibly within its training distribution but does not stress-test the player's plan from novel angles. On the planner side, the same interpolation suppresses novel options: hijacked airliners as missiles before September 2001---explicitly named as a ``failure of imagination'' by the official inquiry \citep{kean2004nineelevencommission}---and large-scale drone use in conventional ground war before 2022 \citep{kunertova2023ukrainedrones} are obvious retrospectively but largely absent from prior planning discourse, and the wargames that mattered would have had to invent those options. Recent agent benchmarks document the pattern across both sides: LMs do well on prescriptive tasks where the next move is well-cued by context, and noticeably worse on open-ended planning where the gain from exploration is high \citep{paglieri_balrog_benchmarking_2025, costarelli_gamebench_evaluating_2024, yao_spinbench_how_2025, alyahya_zerosumeval_extensible_2025}. The aspirational use of an AI participant is to be at least as creative as the human participants; the realized capability is closer to a fluent restatement of the strategy literature already in pretraining. The gap between aspirational and realized creativity, on either side, is itself a safety property that should be measured rather than assumed.

These failures compound: fluent but opaque adjudications carry unsupported assumptions into final summaries (decision laundering); role collapse lets the same latent biases produce both the move and the adjudication that justifies it; an adjudicator's early framing hardens into durable simulated reality across turns; and weak strategic imagination leaves the exercise internally fluent but strategically narrow. None of this is to claim that LMs are useless in every adjudication role; tool-grounded adjudication of bounded, rule-rich subdomains can be reliable enough to use with appropriate scaffolding \citep{zeng_setting_dc_2025}. These are not new failures of LMs as such; they are failures of the role structure an LM is asked to play in an open-ended wargame. The safety question is therefore not, ``Did the agent win?'' It is, ``Which generated claims entered the simulated world, why were they accepted, and what real decisions might they influence?''

%% file: sections/11-app-evaluation-safetycase.tex
\section{Evaluation and Safety-Case Evidence}
\label{sec:app-evaluation-safetycase}
\label{sec:safety-cases-needed}

\subsection{Expanded benchmark critique}

\input{sections/11-app-evaluation-safetycase/original-benchmarks}

\subsection{Expanded safety-case argument}

\input{sections/11-app-evaluation-safetycase/original-safetycase-pre}

\input{tables/tab-safetycase}

\input{sections/11-app-evaluation-safetycase/original-safetycase-post}

\subsection{Compact checklist for authors and practitioners}

\begin{itemize}
  \item State the exercise purpose, decision context, and forbidden uses.
  \item Identify every human and model role: player, adversary, adjudicator, facilitator, analyst, summarizer, and reviewer.
  \item Record model provider, model version, decoding settings, tools, retrieval sources, prompts, and system instructions.
  \item Prevent unreviewed role collapse: the same model should not propose, adjudicate, and summarize a high-impact claim without independent review.
  \item Capture transcripts, moves, adjudications, scenario-state updates, dissent notes, and final summaries.
  \item Require adjudicators to state causal assumptions, evidence, uncertainty, and rejected alternatives for major world-state changes.
  \item Run paraphrase, prompt-sensitivity, cross-model, and red-team robustness tests.
  \item Include escalation and de-escalation stress tests when the scenario involves conflict, coercion, cyber operations, public health, or crisis response.
  \item Train stakeholders on LM failure modes, including sycophancy, hallucination, long-context drift, and narrative overconfidence.
  \item Audit after-action products for unsupported claims, overconfident language, and decision laundering.
\end{itemize}

%% file: sections/11-app-evaluation-safetycase/original-benchmarks.tex
Benchmarks remain useful as intermediate stepping stones (they can isolate hallucination, sycophancy, adjudication errors, or rule adherence under monitored conditions), but they are not sufficient for high-stakes open-ended wargames, for three related reasons \citep{kapoor2024agents}. First, \emph{benchmark over-optimization}: once the test set is known, models can be tuned along the dimensions the benchmark captures while leaving everything else underdeveloped, including the safety properties this paper is most concerned with. Second, distributional mismatch: a benchmark assumes a reasonably stable task distribution, specified inputs, repeatable outputs, and criteria for success; open-ended wargames intentionally violate these assumptions. The action space is not fixed; players adapt to each other; objectives evolve; adjudication involves judgment; and the value of the exercise frequently lies in surfacing assumptions rather than finding an optimal move. Third, ground-truth absence: the outputs of serious wargames are speculative futures, against which there is no observable correct answer; quality has to be evaluated through expert qualitative review of the exercise traces, not by comparing against a held-out label set. Benchmarks are therefore one component of a safety case, not the safety case itself.

This does not mean open-ended wargame evaluation is impossible. It means the field needs a different evidentiary standard, closer to \emph{measurement validity} than to benchmark performance \citep{jacobs2021measurement}. Closed-game benchmarks answer capability questions: can the model plan, negotiate, deceive, cooperate, or follow rules? Open-ended wargame safety asks additional questions tied to role boundaries, disclosed uncertainty, explicit causal assumptions, prompt-induced strategic drift, and limits on presenting speculative scenario branches as validated conclusions. These are not reducible to win rate.

LM-as-judge methods are also inadequate. Judge models introduce systematic distortions, prompt sensitivity, and correlated errors \citep{li_llmsasjudges_comprehensive_2024}; in open-ended settings, an LM judge may share the blind spots of the LM player or adjudicator and reward narrative plausibility over strategic validity. A safety case must show that the judge does not collapse into family-bias with the player or adjudicator, and must distinguish what the model generated from what was independently verified: evidence that ultimately requires expensive human SME adjudication, because tacit expertise and contested assumptions are often the point of serious wargaming.

Human-likeness is also an insufficient target. In agent-to-agent settings, strategic equilibria may differ from human behavior: LM agents can coordinate, collude, or exhibit sycophancy in ways humans would not \citep{askell_general_language_2021, sharma_understanding_sycophancy_2024, agrawal_evaluating_llm_2025}. A safe adversary model need not be maximally human-like; it must be calibrated, role-faithful, non-sycophantic, and clear about uncertainty: properties closer to role separation, robustness, and uncertainty than to a Turing-style benchmark.

Explainability and actionability also need a domain shift. Ordinary agent benchmarks ask why the model selected an action; for open-ended wargames, the important object is the \emph{adjudication trace}: why a contested action was accepted, what assumptions linked action to consequence, what alternatives were considered, and how uncertainty was represented. The adjudication trace is the primary safety-case evidence for the paper's adjudication and human-oversight components: what a reviewer or facilitator inspects to decide whether a model output should influence a downstream judgment. Mechanistic-interpretability tools \citep{huben_sparse_autoencoders_2024, chen_persona_vectors_2025, tan_analysing_generalisation_2024} may help inspect persona stability or role leakage, but even perfect mechanistic interpretability cannot replace exercise-level traces in a sociotechnical system that couples model behavior, scenario design, facilitation, incentives, and after-action interpretation.

The field therefore needs a standard that treats metrics as components, not substitutes, for safety evidence. That standard is a safety case.

%% file: sections/11-app-evaluation-safetycase/original-safetycase-pre.tex
A \emph{safety case} is a systematic argument, supported by evidence, that a system is acceptably safe for a specified use \citep{kelly1998arguing}, recently adapted to advanced AI systems by \citet{clymer2024safetycases}. The core idea is not new to safety engineering---it is the basis of Goal Structuring Notation---but it is underused in LM wargaming. The key difference from a benchmark is scope. A benchmark reports performance under a task distribution. A safety case says what the system is for, what hazards matter, what controls are in place, what evidence supports those controls, what counter-evidence the argument must address, and what conclusions users are allowed to draw.

For open-ended wargames, the central norm should be simple: \textbf{no safety case, no high-stakes use}. A safety case does not certify that a wargame is true. It documents why a particular LM-enabled exercise is, or is not, fit for a particular decision-support role. \reftab{safetycase} sketches the argument structure.

%% file: tables/tab-safetycase.tex
\begin{paperwidetable}[t]
\caption{Argument-structure skeleton for a high-stakes LM-enabled open-ended wargame safety case, following safety-engineering practice \citep{kelly1998arguing}. \textbf{Top claim:} \emph{This LM-enabled wargame is acceptably safe for the specified decision-support role,} decomposed into sub-claims with the evidence each requires and the counter-evidence it must address.}
\label{tab:safetycase}
\centering
\scriptsize
\setlength{\tabcolsep}{4pt}
\renewcommand{\arraystretch}{1.1}
\begin{tabularx}{\linewidth}{>{\raggedright\arraybackslash}p{4cm}YY}
\toprule
\textbf{Sub-claim} & \textbf{Supporting evidence} & \textbf{Counter-evidence the sub-claim must address} \\
\midrule
1. Outputs reach only the decision contexts named in scope. & Documented decision path; named uses and forbidden uses; audit of consumers of after-action products. & Outputs reaching contexts not in scope; reuse beyond stated decision-support role. \\
2. No model instance proposes, adjudicates, and validates the same high-impact claim. & Role map; per-instance assignment logs; cross-model independent review at decision points. & Same model instance crossing roles; correlated errors across instances sharing training data or prompts. \\
3. Major world-state changes are grounded in stated assumptions, sources, and alternatives. & Per-change adjudication trace: causal assumptions, retrieved sources, alternatives considered, uncertainty stated. & Opaque adjudications; ``the model decided'' without rationale; uncertainty hidden in fluency. \\
4. Behavior is robust to perturbations representative of real wargame inputs. & Paraphrase, prompt-sensitivity, cross-model, and red-team test results on representative move classes. & Fragility under standard adversarial prompts; off-distribution behavior unaccounted for. \\
5. Human SMEs can intervene and override at named decision points. & Documented oversight points; SME training on LM failure modes; recorded human--model disagreements. & Disagreements suppressed in summary; SMEs given purely advisory role; oversight absent at the highest-leverage decisions. \\
6. After-action summaries do not assert more than adjudication traces support. & Post-game audit findings; sentence-level mapping from summary claims to traced adjudications. & Summary asserts conclusions unsupported by the trace; decision laundering through narrative compression. \\
\bottomrule
\end{tabularx}
\end{paperwidetable}

%% file: sections/11-app-evaluation-safetycase/original-safetycase-post.tex
Consider COA-GPT \citep{goecks_generative_2024}, a DEVCOM Army Research Lab GPT-4 system that generates Courses of Action under human-in-the-loop oversight. Mapped to \reftab{safetycase}, its published evaluation supports sub-claim~5 ($\checkmark$; named commander selects and adapts COAs) and partially supports sub-claims 1 and~4 ($\triangle$). It does not appear to address sub-claim~2 ($\times$; no separation between proposer and evaluator), sub-claim~3 ($\times$; no auditable assumption traces), or sub-claim~6 ($\times$; no audit of downstream over-claim). The gap between published evaluation and a deployment-ready safety case is wide, even for systems this far along, and naming what is missing is the work a safety-case norm forces.

The safety-case norm also assigns responsibilities. \textbf{AI researchers} should employ open-ended wargames as stress tests for open-ended, decision-influencing agents rather than treating closed games as sufficient proxies. \textbf{Benchmark designers} should report the player- and adjudicator-side openness of an environment and should avoid claiming open-ended strategic safety from analytical adjudication alone. \textbf{Model developers} should publish wargame-relevant failure profiles: escalation behavior, prompt sensitivity, persona stability, long-context drift, sycophancy, and calibration under strategic uncertainty. \textbf{Interpretability researchers} should study adjudication traces, role leakage, and causal assumptions, not only action selection. \textbf{Practitioners} should require artifact capture and safety cases before using LM-generated wargame outcomes in high-stakes planning. \textbf{Conference reviewers and funders} should reward research that studies when open-ended AI exercises should not be trusted.

The standard scales with consequence: a low-risk educational game may need simply lightweight logging and human review; a business workshop on public information may need model/version logs, assumption tracking, and evidence requirements; a national-security, public-health, or critical-infrastructure exercise should require stronger role separation, SME adjudication, escalation testing, adversarial-prompt evaluation, and post-game audit.

\subsection{Facilitator SME Audit Rubric for Adjudication Traces}

To operationalize safety-case sub-claim~3 (Adjudication Auditability) during live exercises, human facilitators and subject-matter experts (SMEs) should evaluate exercise traces against a 5-point audit rubric:

\begin{enumerate}[leftmargin=1.5em,itemsep=2pt,topsep=3pt]
  \item \textbf{Causal Explicitness:} Are all state changes, success probabilities, and side effects supported by explicit causal reasoning in the trace rather than silent pattern completion?
  \item \textbf{Role Independence:} Is the adjudicator model family isolated from player and adversary model families to prevent self-confirming feedback loops?
  \item \textbf{Escalation Attribution:} Is escalatory scenario momentum directly traceable to explicit player decisions rather than model-introduced hostile bias?
  \item \textbf{Strategic Continuity:} Does scenario state maintain logistics, historical constraints, and force availability consistently across long-context turns?
  \item \textbf{Uncertainty Disclosure:} Does the adjudication trace explicitly flag speculative completions, missing evidence, and low-confidence inferences before updating scenario state?
\end{enumerate}

A safety-case norm also changes what counts as progress. A new model that produces more plausible diplomatic dialogue is not automatically safer. A model that exposes uncertainty, resists role leakage, documents causal assumptions, and declines to adjudicate beyond evidence may be more valuable for high-stakes wargaming than a model that is merely more fluent. Progress means learning where an LM-enabled exercise should not be trusted, not only making the model more fluent.

%% file: sections/12-app-objections.tex
\section{Alternative Views and Objections}
\label{sec:objections}

\input{sections/06-objections/body}

%% file: sections/06-objections/body.tex
\textbf{Objection 1: Wargames are too niche, too military, or too dual-use for ML venues.} Military applications raise dual-use concerns, and the ML community should not accelerate unsafe operational deployment. The response is that open-ended wargaming is broader than the military (\refsec{decision-influence-not-gameplay}), and the safety problem already concerns ML practice. More importantly, the position is not ``deploy LM wargames.'' The position is that if the field builds or studies decision-influencing strategic agents, it should examine the settings in which those agents can shape narratives, adjudicate consequences, and affect human decision. Safety cases can reduce irresponsible use by making limits explicit. Nor do we claim open-ended wargames are the largest AI-risk pathway, only a neglected one given how close they sit to escalation and institutional decision-making.

This paper also separates open-ended wargaming from direct weapon control. Many autonomous or remotely piloted systems involve bounded perception-action loops, even when their deployment raises severe safety and governance concerns. Open-ended wargames, in contrast, operate through natural-language scenario generation, role-play, adjudication, and after-action interpretation. The primary hazard is upstream decision influence: a model changes what humans believe about adversaries, escalation, feasibility, or necessity. The domains can interact---wargames can influence doctrine or operational concepts for autonomous systems---but the safety case for an LM-enabled wargame should not be collapsed into the safety case for a weapon platform.

\textbf{Objection 2: Open-ended wargames are not reproducible enough for scientific evaluation, and closed games would be better science.} These are two sides of the same concern, and we treat them together. Open-ended wargames resist the clean repeatability of closed-game benchmarks. Closed games isolate variables, permit large-scale simulation, and support precise scoring \citep{silver_mastering_game_2017, perolat_mastering_game_2022}, whereas open-ended wargames vary across runs. But reproducibility should not be confused with determinism. Human-subject experiments, qualitative studies, and systems evaluations often study phenomena that are only partially repeatable. Open-ended wargames can improve reproducibility through artifact capture (structured logging of prompts, model outputs, adjudication decisions, and after-action traces), fixed scenario scaffolds (pre-specified vignettes that constrain initial conditions across runs), multiple runs, cross-model comparisons, SME review rubrics (methodical assessment forms used by subject-matter experts to assess traces), and clear logging of adjudication criteria. We are not arguing that closed games should be abandoned; they remain the right setting for many scientific questions about planning, search, and self-play. We are arguing against using closed-game performance as evidence for a stronger claim: that an agent is safe for open-ended strategic decision support. A model that performs well in StarCraft or Diplomacy may still fail when it must adjudicate ambiguous cyber coercion, interpret diplomatic signaling, or summarize a crisis game for decision-makers. If a safety-critical use case is hard to evaluate, avoiding it does not make it safer.

\textbf{Objection 3: LMs are not capable enough yet for this to matter.} Current LMs do fail in wargaming contexts; studies have found brittle reasoning, hallucinations, rule non-adherence, inconsistency, and escalation risks \citep{lamparth_human_vs_2024, rivera_escalation_risks_2024, shrivastava_measuring_freeform_2024}. That is a reason to study safety now, not later. The introduction makes the converse point: as long as LMs appear capable in some prominent setting, some users will find it irresistible to extend that perception into settings where the actual capability is much weaker, even when the belief is misguided. Halo effects from frontier-model demonstrations in adjacent domains can make a brittle wargame model appear ready. The argument is reinforced by the relational character of several failure modes in \refsec{domain-specific-failures}: prompt sensitivity and sycophancy under leading questions are jointly produced by model and user, so the safety case must address the human side as well as the model side. Waiting until models are broadly capable as players and adjudicators would leave the field with capability demonstrations but no norms for auditability, interpretability, or decision influence.

\textbf{Objection 4: Safety cases become bureaucratic checklists with no real enforcement.} A safety case that merely records that a benchmark was run and a human was present degenerates into safety-washing. To avoid bureaucratic compliance, a safety case must be evaluated as a rigorous argument tested by its negative claims: explicit evidence showing where hazards exceed controls, where evidence fails, and where deployment is prohibited. Enforcement rests on existing channels in each context: procurement in defense, reviewer gating and model-card disclosure in academic publication, internal review boards, and post-incident audit. Regulatory enforcement is years away in most jurisdictions; a soft norm carried through these channels is the realistic starting point, not the end state.

\textbf{Objection 5: High-stakes LM-enabled wargaming should not exist; auditing it legitimizes the use.} AI-governance reviewers may argue that interacting with LM-enabled wargames, even to demand auditing, tacitly endorses the practice. Some uses should not exist at all, including real-time crisis-response systems where a model adjudicates consequences for a decision-maker with minutes to act. Our position is consistent: auditability is a precondition for high-stakes use, not a license to deploy. ``No safety case, no high-stakes use'' encompasses ``no valid safety case is currently possible, therefore no deployment'' for un-certifiable roles. Refusing to study how these systems fail forfeits the technical grounding needed to articulate refusal.